%% file: lstm_prediction.tex
\documentclass[letterpaper, 10 pt, conference]{ieeeconf}  

\IEEEoverridecommandlockouts                              

\let\labelindent\relax

\usepackage{graphics} 
\usepackage[pdftex]{graphicx} 
\usepackage{amsmath} 
\usepackage{amssymb}  
\usepackage{blindtext}
\usepackage{todonotes}
\usepackage{acro}
\usepackage[normalem]{ulem}
\useunder{\uline}{\ul}{}
\usepackage{colortbl}
\usepackage{float}
\usepackage{balance}
\usepackage{cleveref}
\usepackage[inline]{enumitem}
\usepackage{siunitx}
\usepackage{mwe,tikz}
\usepackage[percent]{overpic}
\usepackage{subcaption}
\usepackage[font={footnotesize}]{caption}
\usepackage[inline]{enumitem}
\usepackage[noadjust]{cite}
\usepackage{url}

\DeclareAcronym{2d}{short = 2D, long = two-dimensional}
\DeclareAcronym{1d}{short = 1D, long = one-dimensional}
\DeclareAcronym{ros}{short = ROS, long = Robot Operating System}
\DeclareAcronym{cnn}{short = CNN, long = convolutional neural network}
\DeclareAcronym{rnn}{short = RNN, long = recurrent neural network}
\DeclareAcronym{fc}{short = FC, long = fully-connected}
\DeclareAcronym{relu}{short = ReLU, long = Rectified Linear Units}
\DeclareAcronym{dnn}{short = DNN, long = deep neural network}
\DeclareAcronym{uav}{short = UAV, long = unmanned aerial vehicle}
\DeclareAcronym{mpc}{short = MPC, long = model predictive controller}
\DeclareAcronym{dwa}{short = DWA, long = dynamic window approach}
\DeclareAcronym{lstm}{short = LSTM, long = Long-Short Term Memory}
\DeclareAcronym{rvo}{short = RVO, long = Reciprocal Velocity Obstacles}
\DeclareAcronym{gp}{short = GP, long = Gaussian Process}
\DeclareAcronym{irl}{short = IRL, long = Inverse Reinforcement Learning}
\DeclareAcronym{rl}{short = RL, long = Reinforcement Learning}
\DeclareAcronym{ae}{short = AE, long = auto-encoder}
\DeclareAcronym{apg}{short = APG, long = angular pedestrian grid}
\DeclareAcronym{bptt}{short = BPTT, long = backpropagation through time}
\DeclareAcronym{sf}{short = SF, long = social forces}

\newcommand{\mat}[1]{\boldsymbol{\mathbf{#1}}}
\newcommand{\x}{\mat{x}}
\renewcommand{\v}{\mat{v}}
\renewcommand{\u}{\mat{u}}
\newcommand{\X}{\mat{X}}
\renewcommand{\H}{\mat{H}}

\newcommand{\g}{\mat{g}}
\newcommand{\G}{\mat{G}}
\newcommand{\params}{\mat\theta}
\newcommand{\F}{\mathcal{F}}

\usepackage{xcolor}

\title{A Data-driven Model for Interaction-aware Pedestrian Motion Prediction in Object Cluttered Environments}
\author{Mark Pfeiffer, Giuseppe Paolo, Hannes Sommer, Juan Nieto, Roland Siegwart, and Cesar Cadena%
\thanks{This work has received funding from the European Union Seventh Framework Programme FP7, project EUROPA2, Grant No. 610603 and Horizon 2020, project CROWDBOT
Grant No. 779942.}
\thanks{The authors are with the
ETH Zurich, Zurich, Switzerland. \newline
{\tt \{pfmark, giupaolo, sommerh, nietoj, rsiegwart, cesarc\}@ethz.ch}.}%
}

\begin{document}

\maketitle
\thispagestyle{empty}
\pagestyle{empty}

\input{abstract}

\input{introduction}

\input{related_work}

\input{approach}

\input{experiments}

\input{conclusion}

\footnotesize
\bibliographystyle{style/IEEEtran}
\balance
\bibliography{bib/IEEEfull.bib,bib/lstm_prediction}

\end{document}

%% file: abstract.tex
\begin{abstract}
This paper reports on a data-driven, interaction-aware motion prediction approach for pedestrians in environments cluttered with static obstacles.
When navigating in such workspaces shared with humans, robots need accurate motion predictions of the surrounding pedestrians.
Human navigation behavior is mostly influenced by their surrounding pedestrians and by the static obstacles in their vicinity.
In this paper we introduce a new model based on Long-Short Term Memory (LSTM) neural networks, which is able to learn human motion behavior from demonstrated data.
To the best of our knowledge, this is the first approach using LSTMs, that incorporates both static obstacles and surrounding pedestrians for trajectory forecasting.
As part of the model, we introduce a new way of encoding surrounding pedestrians based on a 1d-grid in polar angle space. 
We evaluate the benefit of interaction-aware motion prediction and the added value of incorporating static obstacles on both simulation and real-world datasets by comparing with state-of-the-art approaches.
The results show, that our new approach outperforms the other approaches while being very computationally efficient and that taking into account static obstacles for motion predictions significantly improves the prediction accuracy, especially in cluttered environments.
\end{abstract}

%% file: introduction.tex
\section{Introduction}
\label{sec:introduction}

A vast amount of research has been done in the area of robotic navigation in static and known environments.
Yet, when bringing robots from static or controlled environments to dynamic ones with agents moving in a varied set of patterns, many unsolved challenges arise \cite{trautman2010unfreezing}.
In order to integrate robots in a smooth fashion into the workspace shared with other agents --- in many cases pedestrians --- their behavior needs to be well understood such that accurate predictions of their future actions can be made.
While humans can rely on their ``common sense'' and experience for reading other agents' behavior, robots are restricted to pre-defined models of human behavior and interactions.
Independent of the robotic platform, ranging from small service robots to autonomous cars, high accuracy of the other agents' motion predictions is crucial for efficient and safe motion planning.

The main factors influencing pedestrian motion are interactions among pedestrians, the environment, and the location of their destination.
By taking into account the interaction between pedestrians, the accuracy of the motion models can be significantly increased \cite{pfeiffer2016iros,kretzschmar2016social,alahi2016social,vemula2017modeling,trautman2010unfreezing}.
It was shown that using interaction-aware motion models for dynamic agent prediction in motion planning applications makes robots more predictable for human agents \cite{pfeiffer2016iros} and therefore also more ``socially compliant'' \cite{kretzschmar2014learning}.
Especially for cluttered environments it is also important to model the pedestrians' reactions to static obstacles in their close proximity \cite{pfeiffer2016iros,kretzschmar2016social}.

\begin{figure}[t]
  \centering
  \includegraphics[width=0.99\columnwidth, trim=0 0 0 0, clip]{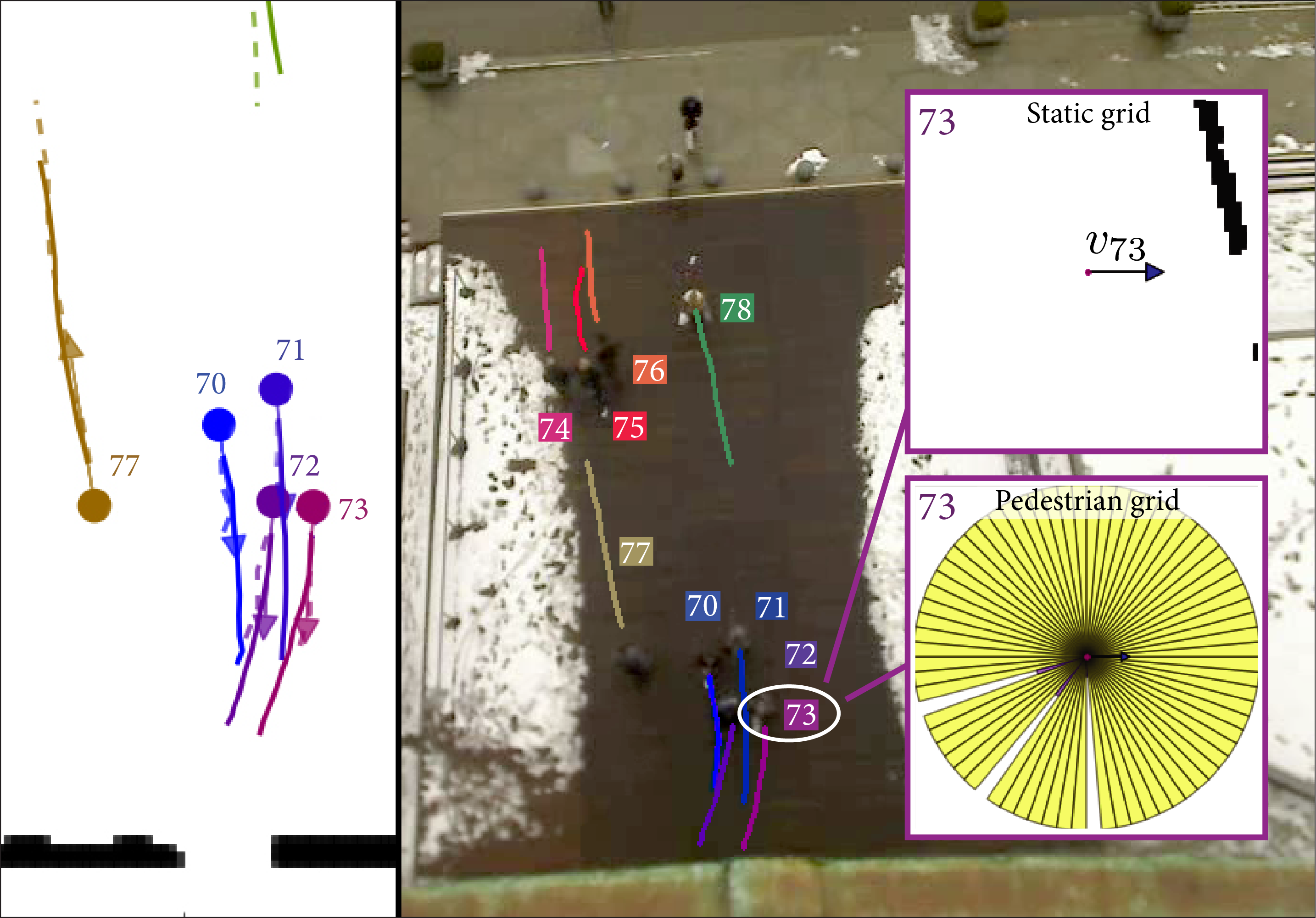}
  \caption{Trajectories of pedestrians predicted with the presented motion model for interacting pedestrians on a real-world dataset.
  \textbf{Left}: Two-dimensional view of the navigating pedestrians, where the ground truth data (dashed) and the predicted trajectories (solid) are shown.
  \textbf{Main image}: Real-world environment with predicted trajectories projected into the image frame.
  \textbf{Right top}: Static obstacle grid of agent 73, which serves as one input to the model. Static obstacles are shown in black.
  \textbf{Right bottom}: Pedestrian grid of agent 73, which is another input to the model. 
  Both grids are aligned with the agent's position and heading, indicated by the arrows in the center of the grids. 
  Each prediction is purely based on such two grids per agent, and its current velocity ($v_{73}$).}
  \label{fig:teaser}
  \vspace{-3mm}
\end{figure}

The existing approaches are limited by at least one of the following shortcomings:
(i) The feature functions, which abstract agent trajectory information to an internal representation, are hand-crafted and therefore can only capture simple interactions. 
(ii) The approaches are not scalable to dense crowds since they use pairwise interactions between all agents \cite{pfeiffer2016iros,trautman2010unfreezing,kretzschmar2016social}, which leads to a quadratic complexity in number of agents, and therefore real-time computation is only feasible for a small number of agents.
(iii) Static obstacles are neglected \cite{alahi2016social,vemula2017modeling,trautman2010unfreezing} and
(iv) knowledge about a set of potential destinations is assumed \cite{trautman2010unfreezing,vemula2017modeling,kretzschmar2016social}.


In this work, we overcome all the above limitations with a novel data-driven approach to model interaction and motion of pedestrians based on a \ac{rnn} \cite{goodfellow2016deeplearning}.
We treat the pedestrian prediction problem as a sequence modeling task; a research field where \ac{rnn}s --- especially \ac{lstm} neural networks \cite{hochreiter1997lstm} --- have shown an outstanding performance within the last five years \cite{graves2014speech,bahdanau2014translation}.
To this end, we present a novel model architecture, which fuses three channels of information per pedestrian:
The pedestrian's current velocity, information about the static obstacles around the pedestrian, and information about surrounding pedestrians.
Given these inputs, the pedestrian's future motion is predicted in a receding horizon manner.
Compared to other approaches, which do forward simulation through repetitive use of 1-step predictions provided by the model \cite{alahi2016social,vemula2017modeling}, we already do the forecast for the whole prediction horizon, and therefore do not need a forward simulation.
One possibility to deploy the presented model on a robotic platform is to predict trajectories of surrounding pedestrians and plan a collision-free path for the robot given the predicted trajectories. 

With the presented architecture, abstract feature representations of the inputs can be learned from the demonstration data instead of handcrafting them.
By using fixed size grid-based input formats, the evaluation time of the model is constant for a varying number of surrounding agents. 
This significantly benefits the approach regarding its scalability to more agents.

The main contribution of this paper is the introduction of a new model architecture based on \ac{lstm} neural networks for pedestrian trajectory prediction that does not suffer from the shortcomings (i)-(iv).
Another major contribution is the way of encoding the information about the surrounding pedestrians in a specialized \acl{1d} representation, which combines low dimensionality with a large infromation content.
We provide an extensive evaluation and comparison against state-of-the-art approaches, both on simulated and real-world data, where we resort to a well-known publicly available dataset for pedestrian motion in public areas \cite{pellegrini2009dataset}.
We hypothesize that by taking into account both pedestrian interactions and the static obstacles, the state-of-the-art prediction approaches can be outperformed.

This paper is structured as follows: Section \ref{sec:related_work} gives an overview of the related work.
Section \ref{sec:approach} introduces the specific problem and our approach to solve it.
Section \ref{sec:experiments} shows our experimental results before we discuss and conclude in Section \ref{sec:conclusion}.

%% file: related_work.tex
\section{Related work}
\label{sec:related_work}

One of the first approaches to model interactions among pedestrians is the social forces model presented by Helbing et al.~\cite{helbing1995social}.
Their model is based on a potential field, or more specifically, attractive forces to model goal-driven behavior and repulsive forces to model obstacle avoidance and interactions among various agents.
Later on, Helbing et al.~adapted the model to interacting vehicles and traffic dynamics \cite{helbing1998generalized}.
Similarly, Treuille et al.~\cite{treuille2006continuum} use continuum dynamics to model naturally moving crowds.
When using a good set of parameters, oscillations in the motion can be avoided \cite{kretz2015oscillations} and they are reliable models for simulation purposes.
However, they cannot be applied to motion forecasting without the knowledge of \emph{all} agents' destinations.

Another well known approach for agent interaction forecasting of holonomic agents is the \ac{rvo} method introduced by van den Berg et al.~\cite{van2008reciprocal}.
The main advantages of the \ac{rvo} approach are its computational efficiency and the guarantee for collision free paths of the agents which not only makes it applicable for prediction but also motion planning applications.
The idea behind \ac{rvo} is to compute a joint set of collision-free velocities based on the assumption of other agents moving with a constant velocity or of a joint collision avoidance effort among all agents.
The original approach was extended by Alonso-Mora et al.~\cite{alonso2013optimal} to non-holonomic platforms.
However, by design, the \ac{rvo} approach requires the knowledge of all agents' target velocities and is only a deterministic model.


When it comes to learning-based interaction modeling approaches, various techniques have been analyzed.
Trautman et al.~\cite{trautman2010unfreezing} formalized the so called ``freezing robot problem'' and pointed out why interaction modeling is an important factor for robot navigation.
They proposed a model based on interacting \ac{gp}s, where an interaction potential combines multiple trajectories with each other; each one described by a \ac{gp} itself.
In order to take into account multimodal behavior for motion planning in human crowds, mixture models were introduced later on \cite{trautman2013multi}.
Another learning based interaction model was presented by Vemula et al.~\cite{vemula2017modeling}, where a \ac{gp} model is used to forecast future agent velocities and destinations.
The predictions are made based on a grid-based representation of the world which collects information about pedestrian positions and heading in the agent's surroundings.
Yet a prior knowledge of destinations is required and no static obstacles are taken into account.

A significant amount of work has gone into modeling interacting pedestrians using maximum entropy \ac{irl}.
Kuderer, Kretzschmar et al.~\cite{kretzschmar2014learning,kretzschmar2016social,kuderer2012rss} introduced a maximum entropy probability distribution for a joint set of continuous state-spaces, also inspired by Ziebart et al.~\cite{ziebart2008maximum} and Henry et al.~\cite{henry2010learning}.
The model is based on a set of hand-crafted feature functions capturing the interaction and collision avoidance behavior of pedestrians.
In our previous work~\cite{pfeiffer2016iros} we extended this approach to an interaction-aware motion planner and showcased the applicability of this approach in real-world experiments.
Since the presented maximum entropy \ac{irl} models rely on sampling methods for expectation value computation during training, a common drawback is the training time.
Wulfmeier et al.~\cite{wulfmeier2017large} proposed an approach based on deep maximum entropy \ac{irl}, where a traversability map of the environment can be learned from demonstration data.
No hand-crafted features are required, yet dynamic obstacles --- such as other pedestrians --- are not taken into account explicitly.
Another approach addressing socially aware motion planning was presented by Chen et al.~\cite{chen2017socialdrl}.
They use deep \ac{rl} to learn motion planning policies instead of using imitation learning \cite{pfeiffer2017perception} of human or other expert demonstrations.
In addition, they show how to induce social norms into the model, rather than learning them.

Alahi et al.~\cite{alahi2016social} introduced the \textit{Social-\ac{lstm}} framework.
They use one \ac{lstm} neural network per agent and pool the information of other agents into a grid-based structure to capture the interactions between multiple agents.
Two versions are introduced: one, where only agent positions are pooled and the other, where complete hidden states of close-by agents' \ac{lstm} models are pooled.
Fernando et al.~\cite{fernando2017attention} introduced another \ac{lstm} interaction model based on attention techniques.
However, the number of agents that can be dealt with specifically is encoded in the architecture of the model.
Both approaches based on \ac{lstm} neural networks only take into account dynamic agents and do not reason about the static obstacle avoidance.
This limits the performance of the prediction models especially confined spaces with many obstacles.

Regarding sequence prediction, \ac{rnn} models --- including \acp{lstm} \cite{hochreiter1997lstm} and Gated Recurrent Units (GRU) \cite{chung2014gru} --- have successfully shown their performance in various applications in the last decade.
They were used to forecast video frames \cite{ranzato2014video}, recognize speech \cite{graves2014speech} or for text translation \cite{bahdanau2014translation}.
In addition, Graves et al.~\cite{graves2013generating} showed that \ac{rnn} models can be used for sequence generation which is also our targeted application.

%% file: approach.tex
\section{Approach}
\label{sec:approach}

This section describes the underlying problem of pedestrian prediction and our approach to solve it.

\subsection{Problem formulation}
\label{sec:problem_formulation}
When navigating in cluttered environments, with other agents and / or static obstacles, humans show an outstanding capability to ``read'' the intentions of other people, perceive the environment and extract relevant information from it.
Ultimately, we want robots to be well integrated into the environment in a socially compliant way such that they can safely navigate, do not get stuck or disturb other traffic participants --- and that at a human like level.
In order to reach this level of autonomy, robots need to have accurate models to forecast the evolution of the environment, including other agents.
By  taking into account the motion predictions of other agents, safe paths can be planned for the robot.
The more accurate the motion predictions of the others, the more predictable will the robot be as the re-planning effort per timestep can be minimized \cite{pfeiffer2016iros}.

In this paper we address the problem of finding a good motion and interaction model for pedestrians from recorded demonstrations such that it can be used for the prediction of pedestrians in motion planning applications.
We want to find a model
\begin{equation}
(\x_{i,t+1:t+K_H}, \H_{t+1}) = \F_{\params} \big ( \X_{t}, \G_{t}, \H_t, i \big ),
\end{equation}
that outputs predictions for the future states $\x_{i,t+1:t+K_H}$ of agent $i$ with a prediction horizon $K_H$.
The current state tuple 
\begin{equation}
\label{eq:Xt}
\X_{t} := [\x_{1,t}, \dots, \x_{i,t}, \dots, \x_{N,t}]
\end{equation}
comprises the state of all relevant $N$ pedestrians (within a certain distance from the pedestrian $i$) at time $t$.
$\G_{t}$ defines an occupancy grid containing information about the static obstacles encoded in a \ac{2d} grid.
$\H_t$ defines a hidden state of the model which propagates information through time and is updated at every timestep.

Since in a real-world application the destinations for other pedestrians cannot be assumed to be known, our model must not rely on a predefined set of target positions but only forecast the future motion of pedestrians according to the perceived data.
Only the current information, i.e. at time $t$, about static obstacles and other agents, together with the hidden state $\H_t$ is used for forecasting the future trajectories.
Importantly, in order to be applicable to a wide range of situations and environments, overfitting to a certain environment needs to be avoided.

\subsection{Motion and interaction model}
\label{sec:model}
Pedestrian trajectory forecasting can be seen as a sequence modeling task.
In order to model the complex associations between agent state, static and dynamic obstacles (i.e. pedestrians) over time, we resort to \ac{lstm} neural networks \cite{hochreiter1997lstm} that already successfully showed their high performance in sequence modeling and prediction tasks in the recent years.

With the model, we want to predict the future motion of the $i$th pedestrian, which in the following will be referred to as the \emph{query agent}.
If path predictions for all pedestrians are desired, the model has to be evaluated once for each pedestrian.
By doing so, the time complexity of the prediction problem for all $N$ pedestrians is $\mathcal{O}(N)$.

\begin{figure}[htbp]
  \centering
  \includegraphics[width=\columnwidth, trim=0 0 0 0, clip]{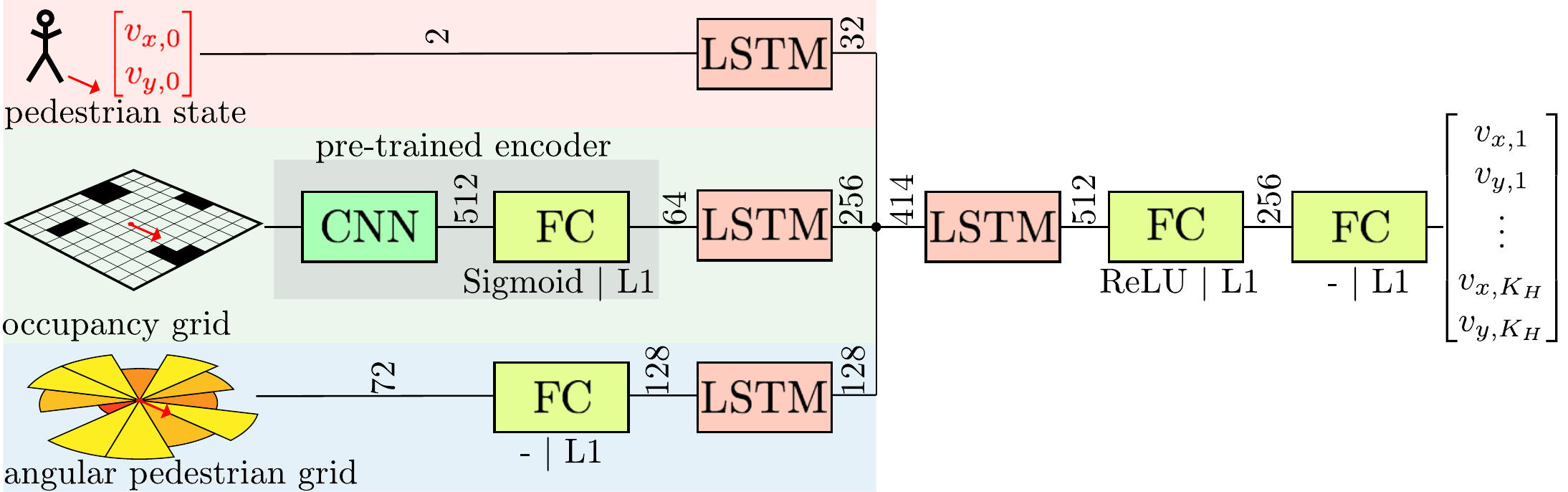}
  \caption{Architecture of the \ac{lstm} model.
  The inputs are the query agent's state (velocity), the occupancy grid and the angular pedestrian grid (APG), all centered at the query agent's position and aligned with its coordinate frame.
  Each of the three different input channels gets processed separately through embedding, \ac{cnn} and \ac{lstm} layers.
  The \ac{cnn} / \ac{fc} combination which pre-processes the occupancy grid is pre-trained with an auto-encoder.
  The concatenation of the extracted features from each channel is followed by another \ac{lstm} layer, a \ac{fc} and an output layer.
  The output of the model is a sequence of velocities predicted for the query agent's future.}
  \label{fig:lstm_model}
  \vspace{-6mm}
\end{figure}

In our model, three channels of information are taken as inputs and fused by a joint \ac{lstm} in order to predict the future behavior of the query agent, as depicted in Figure~\ref{fig:lstm_model}.
The first input, the state, is the query agent's velocity in its local Cartesian coordinate frame $(v_x, v_y)$.
Since all inputs are centered at the query agent's position and aligned with its heading, no position or heading information is required for the presented model.
As pedestrians are holonomic agents, their velocity can differ from their heading.
Therefore, both $v_x$ and $v_y$ need to be taken into account.
It is assumed, that a pedestrian tracker is able to provide the positions and velocities of surrounding agents.

The second input is a \ac{2d} occupancy grid, encoding information about the static obstacles in the vicinity of the query agent.
As a model input we do not use the full occupancy grid $\G_{t}$ but a local extract of it centered at the query agent's position and aligned with its current heading.
When deployed on a mobile robot, such an occupancy grid can be typically obtained based on range finder data.
For this work, it is assumed to be provided.

\begin{figure}[t]
  \centering
  \includegraphics[width=0.7\columnwidth, trim=0 0 0 0, clip]{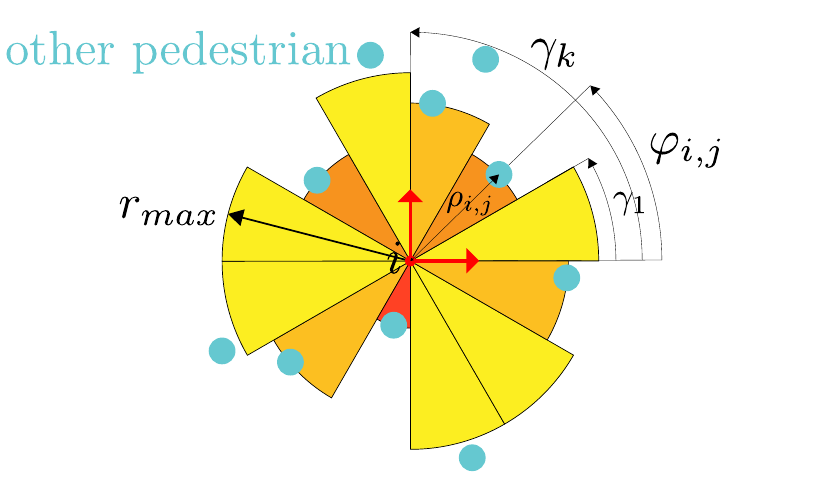}
  \caption{Angular pedestrian grid (APG) construction from relative pedestrian positions ($\X_t$).
  The APG is centered at the query agent's position and aligned with its heading.
  The encoded value per grid cell is the minimum distance to the pedestrians that lie within the cone of the angular grid cell.}
  \label{fig:apg}
  \vspace{-3mm}
\end{figure}

Third, there is the information about the other agents surrounding the query agent, encoded in a special hybrid grid.
In order to be able to capture well the dynamics of other pedestrians, a high resolution grid would be needed, increasing the dimensionality a lot.
Therefore, we introduce a new way of encoding the information of other pedestrian agents with our \ac{apg}, as shown in Figure~\ref{fig:apg}.
The output of the \ac{apg} encoding is a one dimensional vector, $\mat{r}_i$, where each element $r_{i,k}$ is defined by:
\begin{equation}
\begin{split}
r_{i, k} := \min \Big( r_\text{max}, \min \big(\{ \rho_{i,j} \, | j\in \mathcal{N}_{i, k} \} \big) \Big) \\
\mathcal{N}_{i,k} := \big\{ j \in \{1, \dots, N\} \setminus \{i\} | \varphi_{i, j} \in [\gamma_k, \gamma_{k+1}|\big\}
\end{split}
\end{equation}
and $k \in\{ 1,\dots,K\}$, where $K$ defines the number of uniform angular cells, i.e. $\gamma_k := (k - 1) \frac{2 \cdot \pi}{K}$.
The $(\rho_{i, j}, \varphi_{i, j})$ represents the polar coordinates of pedestrian $j$ in the frame of the query agent, $i$, at the query time $t$.
Therefore $\mat{r}_i$ is a function of $\X_t$, $\eqref{eq:Xt}$, only.

Compared to a standard \ac{2d} grid, the resolution of the \ac{apg} only influences the dimensionality of the input linearly, while still being able to capture radial distance changes with continuous resolution instead of discretized grid cells.
In addition, the observability of angular position changes of surrounding pedestrians becomes more precise the closer they are to the query agent.
One drawback of the \ac{apg} representation is the fact that only the closest surrounding pedestrian in each angular cone can be captured, yet we assume that these are the ones that affect the query agents decisions the most.
As in the occupancy grid case, the \ac{apg} is centered and aligned with the pedestrian's position and heading, respectively.

The three inputs to the model are processed as shown in Figure~\ref{fig:lstm_model}.
They get processed separately before being fused into a common \ac{lstm} layer.
The architecture with three separate channels turned out to be more efficient in training than with two big consecutive \ac{lstm} blocks instead.
Two \ac{fc} layers reason about the future velocities, given the output of the common \ac{lstm} cell.
The output of the model is a series of future velocity pairs ($v_x, v_y$) of length $K_H$.
By predicting multiple future steps at once, forward simulation and re-evaluation of the model is not needed and computational efficiency during deployment is improved.
The predicted path for the query agent can be found via Euler forward integration of the velocities from the query agent's position.

\subsubsection{Pre-training of the \ac{cnn}}

\begin{figure}[htbp]
\vspace{-1mm}
  \centering
  \includegraphics[width=\columnwidth, trim=0 0 0 0, clip]{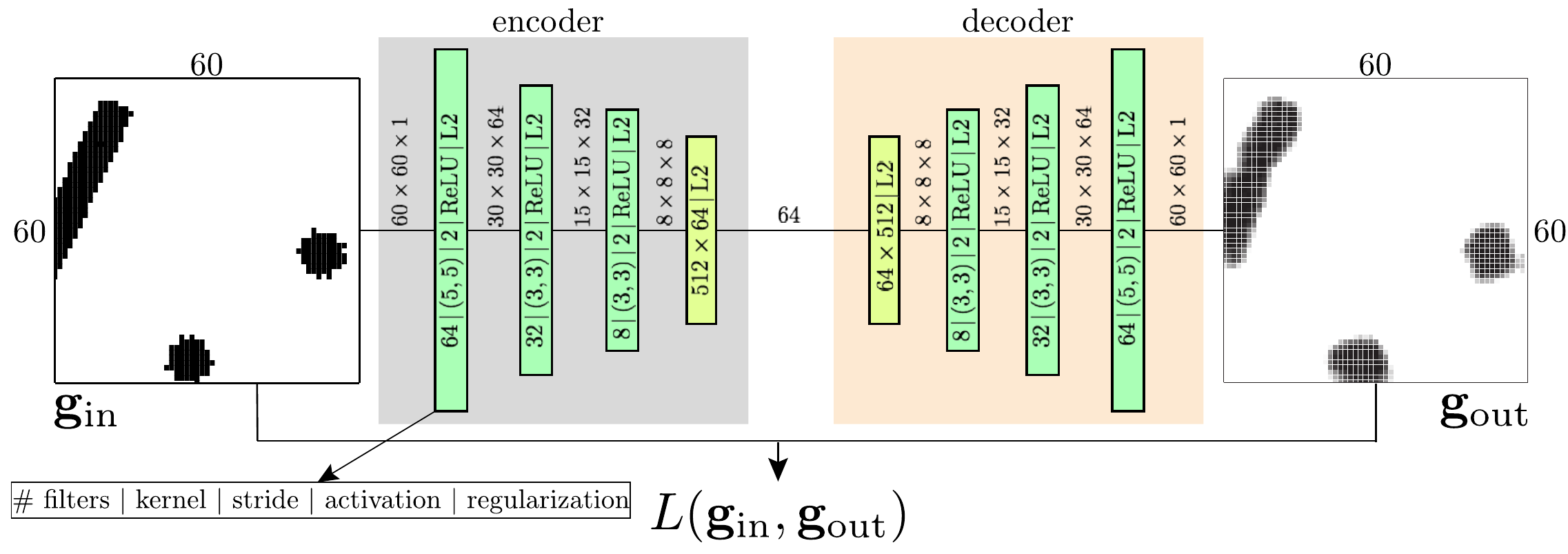}
  \caption{Architecture of the convolutional \ac{ae} network for grid feature extraction.
  By applying three convolutional layers and a \ac{fc} layer, the original occupancy grid ($\g_\text{in}$) gets compressed to a latent space size of 64 during the encoding phase.
  The decoder reconstructs a grid ($\g_\text{out}$) which ideally perfectly matches the input grid.
  The encoding part of this \ac{ae} network is later on used in the full \ac{lstm} motion and interaction model. }
  \label{fig:ae}
  \vspace{-6mm}
\end{figure}

As the extraction of relevant features from the occupancy grid (grey box in Figure~\ref{fig:lstm_model}) is mostly independent of the rest of the overall \ac{lstm} model, we propose to pre-train the \ac{cnn} by using a convolutional \ac{ae} inspired by \cite{masci2011stacked} and \cite{cadena2016multi} as shown in Figure~\ref{fig:ae}.
By doing so, the overall training time can be significantly reduced since the weights of the \ac{cnn} and \ac{fc} encoding are directly affected and not only through backpropagation based on the overall loss.

In the encoding part, the \ac{2d} occupancy grid gets reduced to the latent space of the \ac{ae}.
This \ac{cnn} encoding phase is incorporated into the full \ac{lstm} neural network structure from Figure~\ref{fig:lstm_model}.
The second part of the \ac{ae} is the decoding phase, where the latent space has to be decoded such that the output represents the \ac{2d} occupancy grid.
The \ac{cnn} weight parameters between encoding and decoding are shared and the training error is defined by
\begin{equation}
L(\g_\text{in}, \g_\text{out}) := \sum_{i=1}^{D_x} \sum_{j=1}^{D_y} \big (\g_\text{in}(i,j) - \g_\text{out}(i,j) \big )^2,
\end{equation}
namely the squared error between the encoded-decoded output ($\g_\text{out}$) and the original input grid ($\g_\text{in}$), where $i$ and $j$ define the grid indices along the $x$- and $y$-axis.
$D_x$ and $D_y$ represent the grid dimensions along the axes.
By pre-training the feature extraction for the occupancy grid in this way, it can be assured that the latent space consists of meaningful features, since the original grid needs to be reconstructed from the latent space.

\subsubsection{Model training}
Supervised training is conducted for the motion and interaction model using pedestrian trajectory demonstrations as ground truth.
Using the training data, the model is trained end-to-end, while keeping the \ac{cnn} weights of the encoder (obtained from pre-training) fixed.
The \ac{fc} weights of the encoder are initialized with the values obtained during pre-training but then optimized with the rest of the network.

\begin{figure}[htbp]
\vspace{-2mm}
  \centering
  \includegraphics[width=0.6\columnwidth, trim=0 0 0 0, clip]{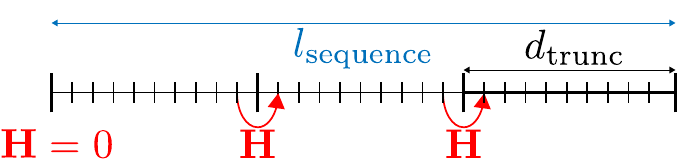}
  \caption{Visualization of the truncated backpropagation through time during training.
  The whole sequence of length $l_\text{sequence}$ gets split up in sub-sequences of length $d_\text{trunc}$.
  For the initial sub-sequence the hidden state of the \ac{lstm}s gets initialized with zeros, for succeeding ones with the last one of the preceding sequence.}
  \label{fig:tbp}
  \vspace{-5mm}
\end{figure}

The full network with its \ac{lstm} cells is trained using \ac{bptt} \cite{werbos1990backpropagation}.
Since the demonstration trajectories have different length and the vanishing / exploding gradient problem \cite{pascanu2013difficulty,bengio1994learning} needs to be avoided, the truncation depth is fixed to $d_\text{trunc}$.
Sub-sequences of length $d_\text{trunc}$ of the demonstrated trajectories are used for training.
For succeeding sub-sequences, the hidden states of the \ac{lstm}s are initialized with the final states of the previous sequence.
In cases when there is no preceding sub-sequence, the hidden states of the \ac{lstm}s are initialized with zeros.
Thus, information from the preceding sub-sequence can be forward propagated to the next one, however the optimization (\ac{bptt}) only affects the samples within the current sub-sequence.
Figure~\ref{fig:tbp} illustrates how the hidden state $\H$ gets propagated between the sub-sequences.

The loss function used for end-to-end training of the model is defined as follows:
\begin{equation}
L (\u_{i}, \params) := \frac{1}{K_H} \sum_{l=1}^{K_H} ||\u_{i}(l) - \v_{\text{gt},i}(l)|| + \lambda \cdot \alpha(\params),
\end{equation}
where $\u_i = [u_i(t+1), \dots, u_i(t+H)]$ represents the predicted velocities by the model for the prediction horizon and $\v_{\text{gt},i}$ the ground truth velocities (from the demonstration data) for the same horizon.
$\alpha(\params)$ introduces the regularization terms and $\lambda$ the regularization factor.
The implemented regularization methods can be found in Figures~\ref{fig:lstm_model} and \ref{fig:ae}.

\subsubsection{Model query during forecasting}
For deployment, at each timestep the current inputs are fed to the network and can be evaluated.
The only difference to a standard feed-forward neural network is the necessity of feeding back the hidden states of the \ac{lstm}s after each query.
Therefore, when a new pedestrian is detected, the hidden state is initialized with zeros, while in later steps the hidden state of the preceding step is used.

%% file: experiments.tex
\section{Experiments}
\label{sec:experiments}
The experiments presented in this section evaluate the added value of an interaction-aware motion model and the new capability to incorporate knowledge of static obstacles into the prediction.

As parameters for the static obstacle occupancy grid, we use a \SI{6}{m} $\times$ \SI{6}{m} grid with a resolution of \SI{0.1}{m} per cell, which leads to a dimension of 60 $\times$ 60 cells.
For the \ac{apg}, we use a maximum range ($r_\text{max}$) of \SI{6}{m} and $72$ ($K$) angular cells, which leads to an angular resolution of \SI{5}{^\circ}.

\subsection{Data assessment}
\label{sec:data_generation}
\begin{figure}[htbp]
\centering
\begin{subfigure}{0.49\columnwidth}
\centering
    \includegraphics[width=\linewidth,trim=8 20 10 20]{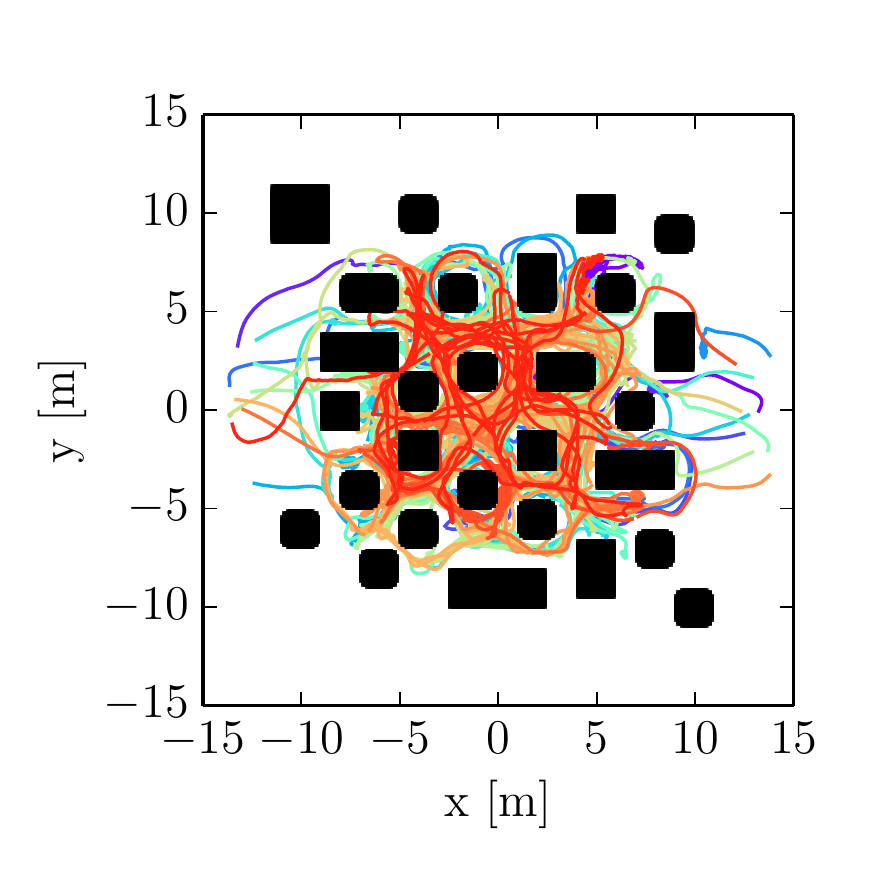}
    \label{fig:training_data_sim}
\end{subfigure}
\begin{subfigure}{0.46\columnwidth}
\centering
    \includegraphics[width=\linewidth,trim=8 20 10 20]{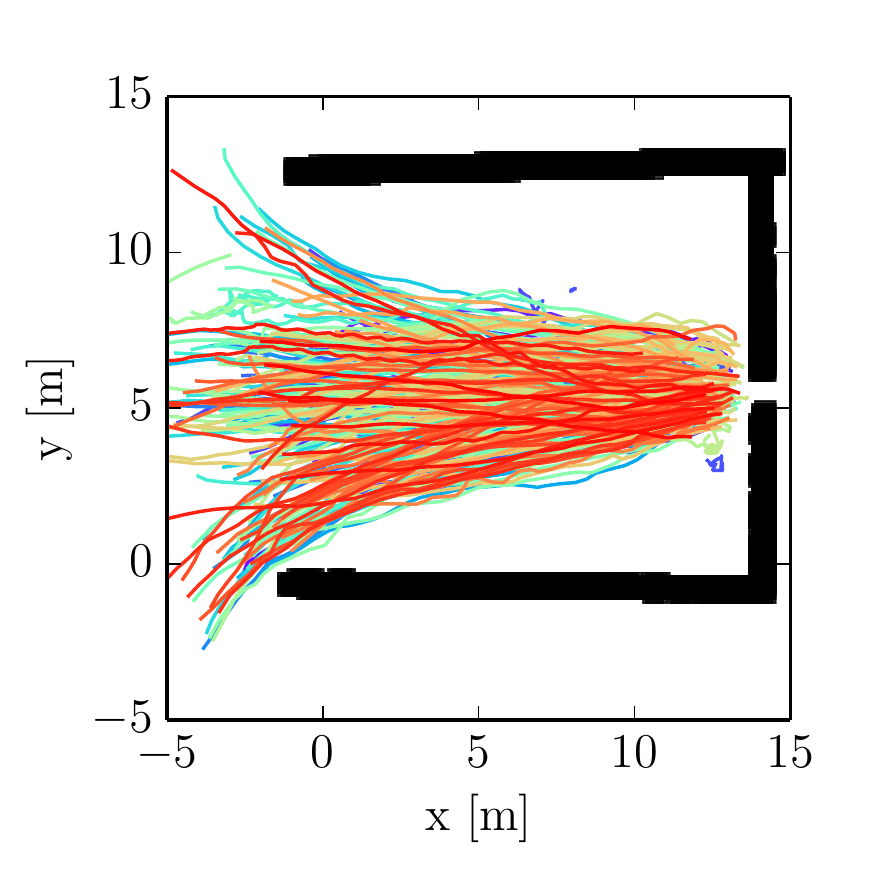}
    \label{fig:training_data_real}
\end{subfigure}
\vspace{-4mm}
\caption{\textbf{Left}: Training data generated with the simulation based on the social forces model \cite{helbing1995social}.
  For better visibility, only a subset of the trajectories is visualized.
  \textbf{Right}: Real-world training data recorded in a public environment \cite{pellegrini2009dataset}.
  The static obstacles are shown in black, the agent's trajectories in colors.}
\label{fig:training_data}
\vspace{-8mm}
\end{figure}

We use both simulated and publicly available real-world \cite{pellegrini2009dataset} datasets to train and evaluate the model.
The simulation data is generated based on the well-known social forces model \cite{helbing1995social} in various environments.
The simulation environments differ in size and the setup of the static obstacles, ranging from an empty corridor to an environment like in Figure \ref{fig:training_data} (left).
In each environment, up to 20 pedestrians are simulated simultaneously while they navigate between randomly changing target positions.
Once a target position is reached, a new one is sampled for each pedestrian, while it is assured that they are not in collision with static obstacles.
As suggested in \cite{helbing1995social}, we add Gaussian noise ($\sigma=0.3$) to the forces generated by the social forces model, which directly affect the acceleration of the pedestrians.
The simulation does not explicitly account for group behavior of walking pedestrians.

\subsection{Evaluation and performance metrics}
\label{sec:performance_metrics}
In order to assess the performance of our model, we compare it against three well known motion prediction approaches. 
Constant velocity (CV) and constant acceleration (CAcc) approaches are the baseline for any motion prediction method.
These models simply assume each agent to continue with it's current velocity or acceleration, respectively.
In addition, we use the social forces and an \ac{lstm} model without knowledge of the static grid for comparison.
For the evaluation on the simulation data we assume the social forces model to significantly outperform all the other approaches used for comparison since the data was generated based on the same model.
The only difference is the added noise on the acceleration.

The \ac{lstm} model unaware of the static grid is similar to the O-\ac{lstm} model, recently presented in \cite{alahi2016social}.
Since the original implementation is not publicly available, we use our own version of the O-\ac{lstm} method for comparison.
Instead of the discrete grid, that solely knows about a pedestrian count per cell, we also use our \ac{apg} to encode the pedestrian information.
This however, contains more detailed information than a standard \ac{2d} grid.
In the following, our introduced model will be referred to as the LSTM model, the one unaware of static obstacles as LSTM-noGrid.

To evaluate the performance of all models quantitatively, we use the prediction error, i.e. the Euclidean distance between prediction and ground truth values, over time.
We analyze the prediction errors for a time horizon of \SI{3}{s}, which requires 10 prediction steps with a sampling time of \SI{300}{ms}.
In addition, we do a qualitative evaluation by visual inspection of the predicted trajectories for multiple examples, both for simulated and real-world data.

\subsection{Simulation results}
\label{sec:sim_results}
Since real-world data oftentimes contains tracking error or pedestrians who are standing still or walking back and forth, we will start our evaluation based on the simulation data.
The simulation results are consistent up to the noise on the accelerations of the pedestrians and therefore are a good way to analyze whether motion policies of pedestrians can be learned by the presented model.
In addition, compared to real data acquisition it is fairly easy to generate new datasets in different environments.

The model tested in simulation is trained in two different environments:
First, it is pre-trained in an empty corridor environment with 10 pedestrians navigating between random target positions at the two sides of the corridor.
With this dataset, the pure interactions with other pedestrians can be learned.
Second, the main part of the training is conducted in the environment shown in Figure~\ref{fig:training_data} (left) and comprises 20 pedestrians.
It turned out that pre-training in a simplistic environment and afterwards continue training in the environment of choice speeds up the overall training time of the model.
The corridor dataset contains less than an hour of data, while the main dataset contains about five hours of navigating pedestrians.
The overall training time is about \SI{3}{h} on a Nvidia GeForce GTX 980 Ti GPU\footnote{https://www.geforce.com/hardware/desktop-gpus/geforce-gtx-980-ti} and requires 300'000 training steps.

\begin{figure}[htbp]
\centering
\begin{subfigure}{0.49\columnwidth}
\centering
    \includegraphics[width=\linewidth,trim=50 100 50 70]{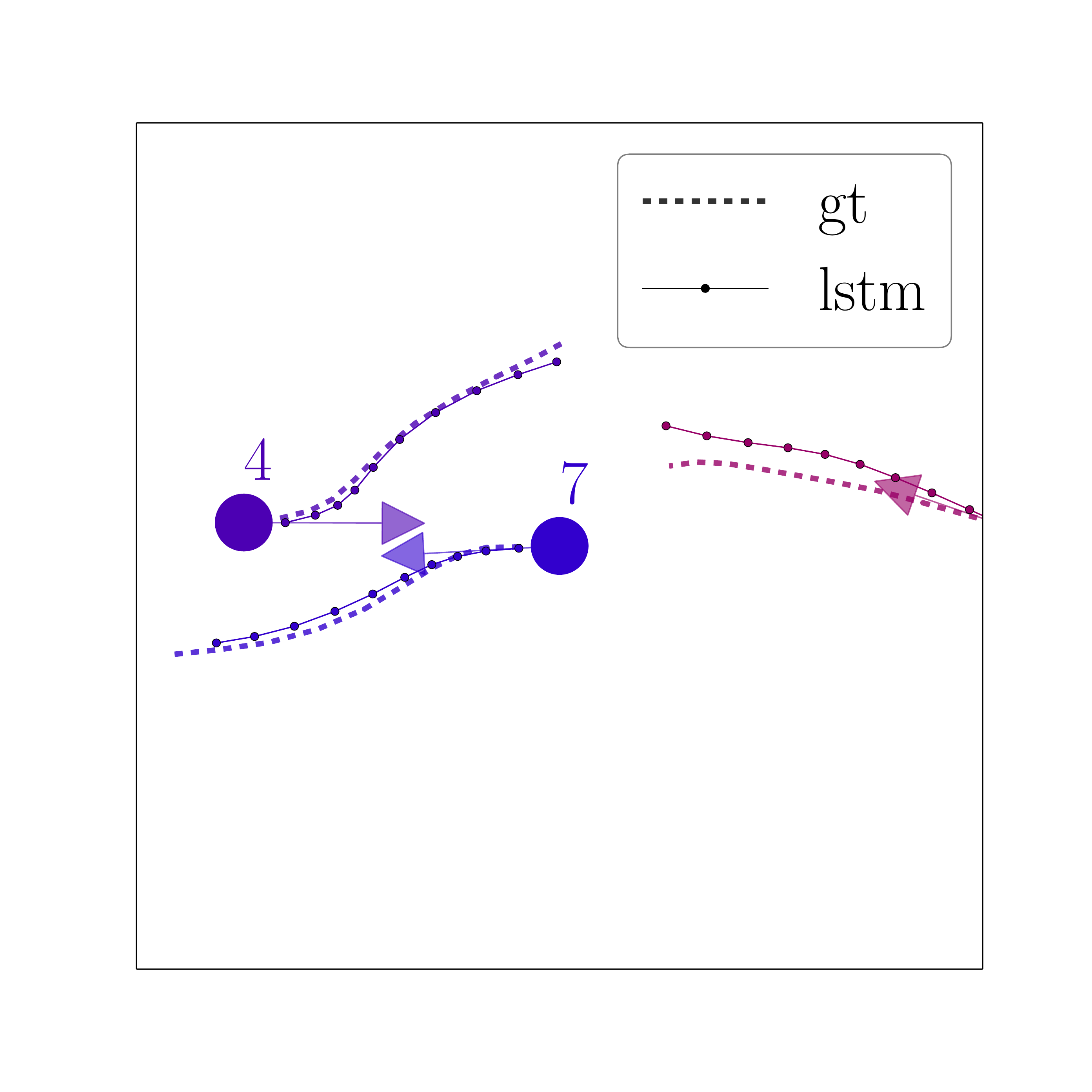}
    \label{fig:traj_examples_1}
\end{subfigure}
\begin{subfigure}{0.49\columnwidth}
\centering
    \includegraphics[width=\linewidth,trim=50 100 50 70]{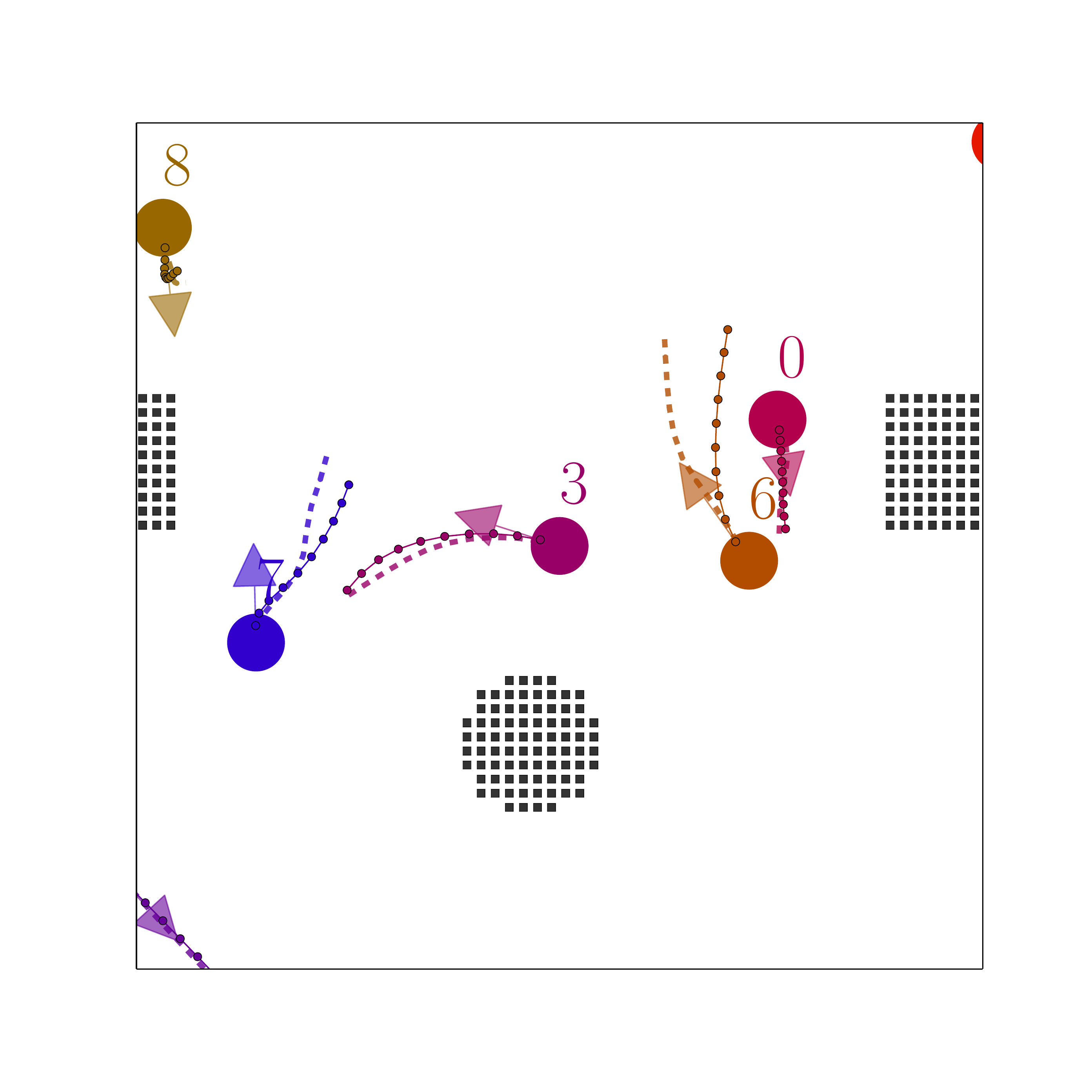}
    \label{fig:traj_example_2}
\end{subfigure}
\vspace{-3mm}
\caption{Two examples for interaction- and obstacle-aware pedestrian trajectory prediction using the presented \ac{lstm} model.
The big circles represent pedestrians (number specifies the agent id), the small ones the ten prediction steps.
The current velocity of each pedestrian is indicated by the arrow.
Static obstacles are shown as black grid cells.
Both examples stem from the test data which was not used for training the model.}
\label{fig:traj_examples}
\vspace{-5mm}
\end{figure}

The evaluation is conducted on a test dataset, which uses a different environment (arrangement of static obstacles) than the one used for training.
Like this, the generalization capabilities of the learned motion- and interaction model can be analyzed.

Figure~\ref{fig:traj_examples} shows two snapshots of the trajectory predictions conducted by the model vs. the ground truth data.
The left image clearly shows how two pedestrians are jointly avoiding each other.
Although the agents' current velocities (arrows) do not show any avoidance measures yet, the \ac{lstm} model already predicts that agent 4 and 7 will both swerve to their left side in order to avoid a potential collision.
The figure also indicates that the predicted avoidance maneuver is similar to the one conducted in the ground truth data, which shows that the model is able to transfer its knowledge between different scenarios.

The right image shows a situation where the future pedestrian trajectories are clearly influenced by the occurrence of static objects.
Although there are no pedestrians in the vicinity of agent 8, our model predicts that this agent will almost come to a stop because of the static obstacle ahead of it.
Agent 7 is predicted to avoid the same static obstacle from the other side while agent 3 needs to do a sharp left turn in order not to collide with agent 7.
Since the model's \ac{lstm} cells provide internal memory, it can keep track of the past motion of an agent.
The internal memory is especially helpful since no target knowledge is available and the predictions only rely on past and current observations.
Compared to the two agents in the left image, where the avoidance effort is shared, agents 0 and 6 in the right image do not share the avoidance.
Agent 6 needs to perform the avoidance maneuver since the static obstacle is blocking agent 0 from moving to its left.
Moreover, the qualitative performance of the predictions can be assessed in the associated video submission\footnote{\url{https://youtu.be/a6ApSImFC3Q}}.

\begin{figure*}[t]
  \begin{subfigure}{\textwidth}
    \includegraphics[width=1\textwidth,trim=10 0 10 5, clip]{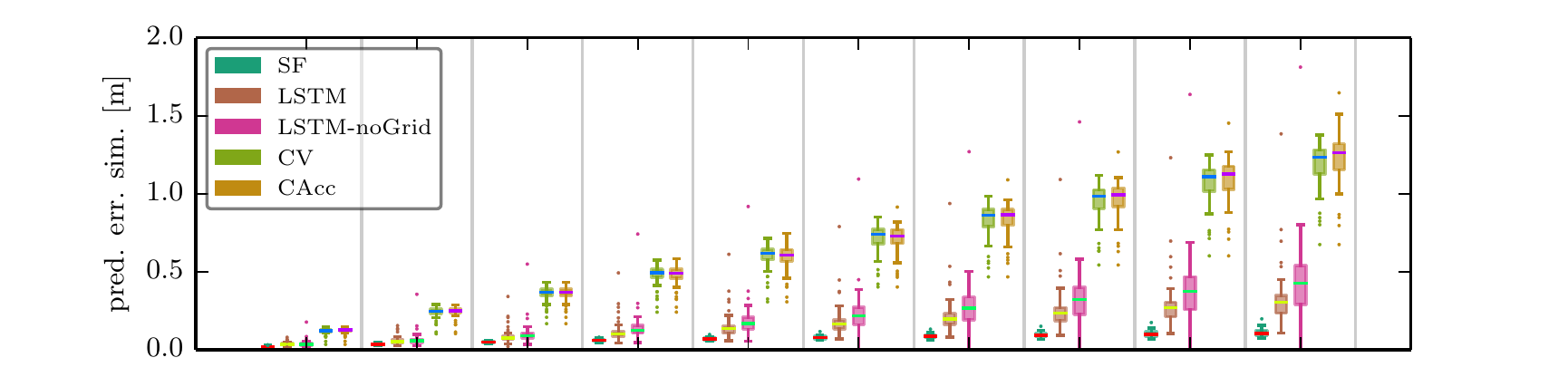}
  \end{subfigure}
  \bigskip
  \begin{subfigure}{\textwidth}
    \includegraphics[width=1\textwidth,trim=10 5 10 5, clip]{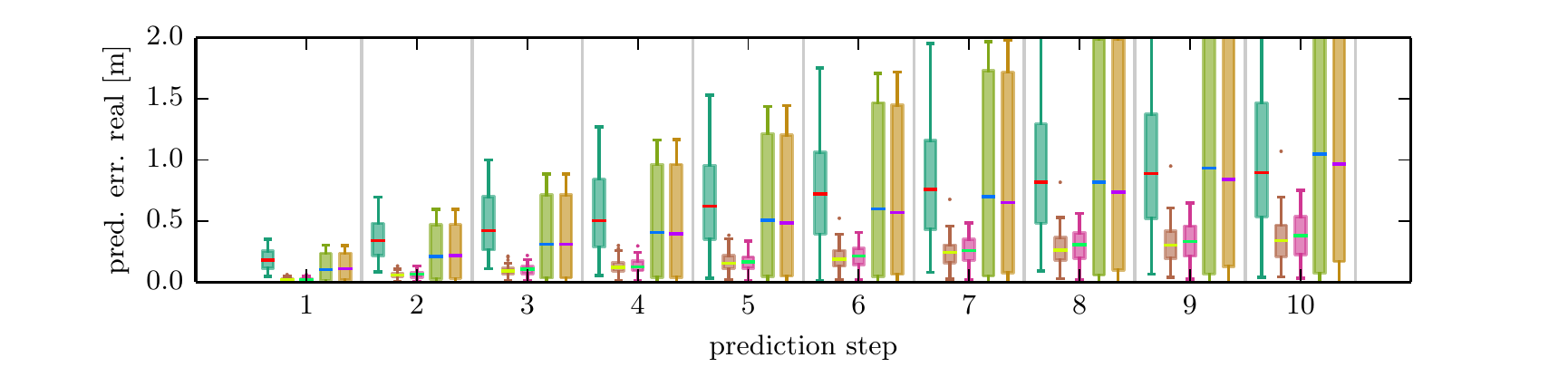}
  \end{subfigure}
  \vspace{-5mm}
\caption{This figure shows the evaluation of the prediction error of various models over a prediction horizon of \SI{3}{s} with 10 prediction steps.
The upper figure presents the prediction error on the synthetic / simulation-based dataset, the lower one the evaluation on the real-world data.
On both datasets, the baseline methods are significantly outperformed by the learning based \ac{lstm} models.
While the simulation data is generated with the social forces (SF) model (and therefore the prediction error is very small), the learning-based \ac{lstm} approaches outperform all the models in the real-world dataset.
The results clearly show that the model taking into account static obstacles (LSTM) significantly outperforms the one unaware of static obstacles (LSTM-noGrid), once the environment is more cluttered.
The presented \ac{lstm} model is not constrained to a single dataset, which is indicated by the similar prediction error for both simulation and real-world dataset.}
\label{fig:boxplot_error}
\vspace{-8mm}
\end{figure*}

The quantitative analysis of the prediction error in the simulation dataset is shown in the upper part of Figure~\ref{fig:boxplot_error} and in Table~\ref{tab:avg_pred_error}.
As expected, the prediction error of the \acl{sf} model is significantly smaller than all other approaches, which stems from the fact that the model was also used for data generation.
Moreover, the \acl{sf} model has perfect knowledge about the destinations of all agents, an element which is not required in the presented \ac{lstm} model.
The prediction error in Figure~\ref{fig:boxplot_error} only stems from the acceleration noise in the \acl{sf} model.
The baseline prediction methods are significantly worse than the \ac{lstm} models, since no knowledge about the environment can be included in the predictions.
Comparing the two \ac{lstm} models, which were both trained on the same dataset, one can see, that adding static obstacles to the input of the model improves its prediction capabilities significantly.
This supports our initial hypothesis.
Especially for a longer prediction horizon, the difference between the model only based on the pedestrian-pedestrian interactions (LSTM-noGrid) and the one using both static obstacles and pedestrians (LSTM) becomes distinct.

Since the inputs to the model are encoded in grids, the evaluation complexity per agent is constant.
This also results in an almost constant (depending on the other processes running on the computer) and predictable query time of the model.
For this example, where 20 pedestrians need to be predicted all the time, the average evaluation time for all agents was \SI{51}{ms} on a standard laptop with an Intel$^{\textregistered}$ Core$\textsuperscript{TM}$ i7-4810MQ CPU processor with 2.80GHz.
This results in an average prediction time of \SI{2.6}{ms} per pedestrian.

\begin{table}[htbp]
\vspace{0mm}
  \centering
  \caption{Average prediction error in meter for the compared models on the simulated and real-world dataset.}
  \label{tab:avg_pred_error}
  \begin{tabular}{c|c|c|c|c|c}
     & \ac{sf} & \ac{lstm} & \ac{lstm}-noGrid & CV & CAcc \\
    \hline
    simulation & 0.071 & 0.169 & 0.220 & 0.652 & 0.659 \\
    real-world & 0.667 & 0.179 & 0.193 & 0.676 & 0.682 \\
  \end{tabular}
  \vspace{2mm}
\end{table}

\subsection{Real-world results}
\label{sec:real_results}
We use a publicly available pedestrian tracking dataset \cite{pellegrini2009dataset} to evaluate our model on real-world data.
Since the real-world dataset is much smaller than the one generated in simulation, it will also give a good indication of how well the model can deal with this amount of information and how prone it is to overfitting.
During training, we use the pre-trained model (trained on the simulation data) and train it with the real-world dataset.

The dataset is recorded at two different locations, both very sparse in terms of static objects.
As for the simulation results, we will use one environment for training while the other will be used for testing only.
The real-world training data is shown in Figure~\ref{fig:training_data} (right).
The dataset also contains information about the group structure of the pedestrian motion.
If pedestrians move close together, they are registered as a group.
We also introduced this information into the \acl{sf} model, as described in \cite{yamaguchi2011you}.
Since the SF model requires destinations per agent we provided it always with the endpoint of the complete observed trajectory as the agent's target.

Figure~\ref{fig:boxplot_error} and Table~\ref{tab:avg_pred_error} show the quantitative analysis of the prediction error in the test environment.
As expected, the constant velocity / acceleration model can only serve as a lower bound for the motion predictions.
The performance of the \acl{sf} model is largely affected by the considerable noise in the real-world data which significantly influences the short-term predictions.
While the prediction error of the other models increases clearly with the horizon, the \acl{sf} prediction error only increases slowly over time which originates from the knowledge of the destinations of the agents.

Since both environments (real-world training and test) barely have any objects blocking the walking areas of the pedestrians, see Figure~\ref{fig:training_data} (right), both the \ac{lstm} and the \ac{lstm}-noGrid model (with and without static objects) perform almost equally in this environment.
However, especially for long term predictions, the prediction errors for the \ac{lstm} model aware of static obstacles is still smaller than the errors of the one unaware of static obstacles.

Compared to the \acl{sf} model, the prediction error can be reduced significantly --- despite the fact that the former has more information available, namely the targets per pedestrian.
Additionally, the prediction errors for the real data are in the same range as for the simulation data.
This also shows that the \ac{lstm} model can learn to deal with noisy input data and can provide reliable interaction-aware predictions in a real-world environment.

A qualitative assessment of the predictions in the real-world environment is shown in Figure~\ref{fig:teaser}.
Although agents 72 and 73 walk straight towards the static obstacle, the model is able to predict that they will walk through the open slot (which actually is a door).
This goes along with the ground truth trajectories.
More examples for the navigation performance in the real-world environment are provided in the associated video submission.

%% file: conclusion.tex
\section{Conclusion}
\label{sec:conclusion}
In this paper, we introduced a new approach to model pedestrian dynamics and interactions among them.
The model architecture is based on an \ac{lstm} neural network which is trained from demonstration data.
To the best of our knowledge, we present the first approach which is able to predict pedestrian-pedestrian interactions and the avoidance of static obstacles at the same time with a neural network model.
In addition, we introduce a new way of handling dynamic objects, the angular pedestrian grid (APG), which encodes the information of the surrounding pedestrians.
In a multi-pedestrian scenario, we use one \ac{lstm} network per pedestrian, and therefore the computational complexity only increases linearly with the number of agents.
This is especially important when the crowd density gets higher.
The presented model does not require a known destination per pedestrian or a predefined set of destination candidates, which makes it well applicable to many real-world applications.

The performance of the interaction- and obstacle-aware motion model is evaluated both on simulation and real-world data.
Our experiments show, that the prediction accuracy of the model can be significantly increased by taking into account static obstacles, especially when the environment becomes more cluttered.

In an environment with only few static obstacles (like the real-world data we evaluated on) our proposed model still clearly outperforms the other, state-of-the-art approaches that we use for comparison.
These include the social forces model, which cannot be applied without providing all agents' target positions and and interaction-aware \ac{lstm} model without static obstacle information.

Additional real-world data needs to be acquired in particular for cluttered environments to further evaluate the performance of our method in different scenarios.
Since the model can make accurate and reliable predictions while being computationally efficient even for a multitude of pedestrians, it is very well suited to be integrated in a mobile platform and to be used for realtime motion planning applications, which will be done in future work.
